%% file: paper-principal-real.tex
\documentclass[preprint]{elsarticle}

\usepackage{hyperref}

\journal{Data Science and Engineering}

\usepackage{amsmath,amssymb,amsfonts}
\usepackage{subcaption}
\usepackage{algorithmic}
\usepackage{graphicx}
\usepackage{ulem}
\usepackage{acronym}
\usepackage[utf8]{inputenc} 
\usepackage[T1]{fontenc}    
\usepackage{url}            
\usepackage{booktabs}       
\usepackage{nicefrac}       
\usepackage{microtype}      
\usepackage{bm}
\usepackage{adjustbox}
\usepackage{multirow}
\usepackage{mathtools}
\usepackage{subcaption}
\usepackage{lineno}
\usepackage{setspace}
\usepackage{enumitem}
\usepackage{comment}
\usepackage{bbm}
\usepackage[table]{xcolor}
\usepackage{caption}
\usepackage{booktabs}

\usepackage{algorithm2e}
\usepackage{subcaption}

\usepackage{ulem}
\usepackage{xcolor}

\acrodef{ML}{Machine Learning}
\acrodef{SSHIBA}{Sparse Semi-supervised Heterogeneous Interbattery Bayesian Analysis}
\acrodef{VAE}{Variational AutoEncoder}
\acrodef{SOTA}{state-of-the-art}
\acrodef{ARD}{Automatic Relevance Determination}
\acrodef{VI}{Variational Inference}
\acrodef{ELBO}{Evidence LowerBOund}
\acrodef{CNN}{Convolutional Neural Network}
\acrodef{MLP}{MultiLayer Perceptron}
\acrodef{BCE}{Binary Cross Entropy}
\acrodef{GLL}{Gaussian Log-Likelihood}
\acrodef{PoE}{Product-of-Experts}
\acrodef{RNN}{Recursive Neural Network}
\acrodef{LSTM}{Long Short-Term Memory}
\acrodef{FA-VAE}{Factor Analysis VAE}
\acrodef{FA}{Factor Analysis}
\acrodef{rv}{random variables}
\RestyleAlgo{ruled}

\include{formulas}

\begin{document}

\begin{frontmatter}

\title{Multi-modal hierarchical Variational AutoEncoders with Factor Analysis latent space}

\author[mymainaddress,secondaryaddress]{Alejandro Guerrero-López\corref{mycorrespondingauthor}}
\ead{alexjorguer@tsc.uc3m.es}
\author[mymainaddress]{Carlos Sevilla-Salcedo}
\author[mymainaddress,secondaryaddress]{Vanessa Gómez-Verdejo}
\author[mymainaddress,secondaryaddress]{Pablo M. Olmos}


\cortext[mycorrespondingauthor]{Corresponding author}

\address[mymainaddress]{Department of Signal Processing and Communications, Universidad Carlos III de Madrid, Leganés, 28911, Spain}
\address[secondaryaddress]{Instituto de Investigación Sanitaria Gregorio Marañón, Madrid, 28007, Spain}

\begin{abstract}
\textbf{Purpose}: Handling heterogeneous and mixed data types has become increasingly critical with the exponential growth in real-world databases. While deep generative models attempt to merge diverse data views into a common latent space, they often sacrifice interpretability, flexibility, and modularity. This study proposes a novel method to address these limitations by combining Variational AutoEncoders (VAEs) with a Factor Analysis latent space (FA-VAE).

\textbf{Methods}: The proposed FA-VAE method employs multiple VAEs to learn a private representation for each heterogeneous data view in a continuous latent space. Information is shared between views using a low-dimensional latent space, generated via a linear projection matrix. This modular design creates a hierarchical dependency between private and shared latent spaces, allowing for the flexible addition of new views and conditioning of pre-trained models.

\textbf{Results}: The FA-VAE approach facilitates cross-generation of data from different domains and enables transfer learning between generative models. This allows for effective integration of information across diverse data views while preserving their distinct characteristics.

\textbf{Conclusions}: By overcoming the limitations of existing methods, the FA-VAE provides a more interpretable, flexible, and modular solution for managing heterogeneous data types. It offers a pathway to more efficient and scalable data-handling strategies, enhancing the potential for cross-domain data synthesis and model transferability.
\end{abstract}


\end{frontmatter}

\newpage
\section{Introduction}

Real-world problems often require the handling of heterogeneous data. For example, combining cloud images with multi-modal information like temperature, humidity, pressure, and wind speed has been shown to improve cloud classification \cite{liu2019hierarchical}. Similarly, fusing multi-modal sensor streams, such as cameras, lidar, radar measurements, and environmental sensors (e.g., temperature, wind speed, and humidity), can enhance object detection for autonomous vehicles \cite{bijelic2020seeing}. Even in microbiology, the combination of spectrograms, epidemiological data, and antibiotic resistance information has been found to optimise the automatic discrimination of bacteria \cite{rodriguez2022rapid, candela2021automatic, guerrero2021development}.

Deep generative models emerge as a powerful methodology for learning from multi-modal data due to their ability to create diverse and realistic synthetic data such as images \cite{hirte2021realistic, li2021single, chen2022two}, text \cite{brown2020language, zheng2021deep, xu2021table}, or audio signals \cite{wu2022wav2clip, borsos2022audiolm, yi2022add}. Deep hierarchical \ac{VAE}s \cite{sonderby2016ladder, klushyn2019learning, vasco2022leveraging, hong2022return} further divide the latent space into disjoint groups of latent variables, enhancing the expressiveness of both the approximate posterior and prior distributions. These models are robust and can disentangle private and shared information into hierarchical latent variables \cite{vowels2020nestedvae, peis2020unsupervised,peis2022missing,kojima2022organization}. However, training such deep hierarchical architectures leads to inference bias due to the use of approximate inference techniques, which often require computationally expensive methods such as Hamiltonian Monte Carlo sampling \cite{peis2022missing, caterini2018hamiltonian}. 

Learning from multi-modal data requires the exploration of complex latent spaces with hierarchical relations, as well as novel design methodologies for incorporating different data modalities or views in a structured manner. In recent years, research has extensively explored such structures. Suzuki et al. \cite{suzuki2016joint} proposed the Joint Multi-Modal VAE (JMVAE), which uses a bidirectional double VAE to project two different views onto a shared latent space. Meanwhile, Li et al. \cite{li2022unsupervised} proposed Joint Variational Adversarial Autoencoder (JVA$^2$E), which combines adversarial learning with the latent features of generative VAE. For example, MVAE \cite{wu2018multimodal} uses separate encoders for $K$ input views and a Gaussian \ac{PoE} to obtain a shared latent representation. This approach is similar to DMVAE \cite{lee2021private}. Another approach is MMVAE \cite{shi2019variational}, which generalises MVAE and DMVAE by replacing the PoE with a Mixture of Experts, resulting in more realistic cross-generated data. Similarly, AMVAE \cite{youpeng2021amvae} projects $K$ views onto a common latent space using a fully connected layer, following a similar idea. For example, Multi-VAE \cite{Xu_2021_ICCV} uses private continuous latent variables to capture view-specific information and categorical discrete latent variables to represent shared information across views. This and other concurrent methods for handling multi-view data share a common aspect: the use of complex hierarchical deep latent representations to capture the correlation among views. However, this complexity brings challenges such as a lack of modularity to incorporate new data views or handle missing views, difficulty in training inference, and limited interpretability regarding the relationship between latent spaces and different views.

In this paper, we present Factor Analysis Variational AutoEncoder (FA-VAE), a flexible and robust two-level hierarchy composed of local per-view VAE-like models combined with a linear FA global latent space, which is capable of effectively handling multi-view data problems. Unlike existing models, FA-VAE offers significant advantages in terms of flexibility, complexity, and modularity. FA-VAE is trained iteratively in two steps. First, each local VAE model handles each heterogeneous view and is updated independently, translating from each domain to a Gaussian-embedded space. Then, the global FA model is updated using a closed-form mean-field formulation given each Gaussian-embedded space. This modular approach allows for pre-training and the incorporation of new views, while also handling missing views naturally. We also propose an innovative approach to enhance interpretability by using \ac{ARD} priors on the FA weight matrices. This allows the analysis of the importance of each global latent dimension in explaining each data view, providing a more nuanced understanding of the underlying data structure. Overall, our approach represents a significant advance in multi-view data modelling, with potential applications across a wide range of domains.

To showcase the capabilities of FA-VAE, we conducted several representative experimental setups. In the first setup, we use a pre-trained unsupervised VAE and condition it to a set of given attributes using FA-VAE with a few mean-field updates. In the second setup, we perform domain adaptation between different image datasets by combining multiple individually pre-trained VAEs. Finally, we demonstrate how FA-VAE can be used to speed up the training of deep complex VAEs by doing transfer learning from a small, simple VAE model over the same dataset. We provide the code to reproduce every experiment on our public repository\footnote{\url{github.com/aguerrerolopez/FA-VAE}}. We believe that our model offers a unique level of flexibility, complexity, and modularity that surpasses existing state-of-the-art models.

\section{Methods}\label{sec:methods}
In the following sections, we propose two different variations of FA-VAE. First, we will analyse the multi-VAE proposal (FA-VAE) where each view is tackled using a marginal $m$-VAE. Then, a second model proposal is detailed, where marginal $m$ -VAEs are combined with non-VAE views (FA-VAE).

\subsection{Factor Analysis Multi-VAE}

The task of multi-view analysis poses a significant challenge in the identification of a common underlying representation for $N$ data samples from $M$ different modalities, denoted as $\llav{\Xnm}_{m=1}^M$, that captures both interview and intraview variability. Achieving this goal requires the fusion of data from multiple modalities into a unified global latent space. However, the latent space must also account for the fact that each view may only be reconstructed by a portion of it or share latent features with a subset of views. 

The primary challenge that needs to be addressed is the handling of heterogeneous data types, such as categorical, image, real, or temporal data. Each modality $\llav{\Xm}_{m=1}^M$ originates from a distinct domain and must be standardised to ensure consistency. In this study, we propose a hierarchical approach that first, given each $\Xm$ employs a marginal \ac{VAE} to convert each $m$-domain into a $m$-Gaussian embedded domain, denoted as $\llav{\mathbf{Z}^{(m)}}_{m=1}^M$. 

The second issue is to consider how to combine this information into a global latent space $\G$. To address this, we propose interconnecting all marginal $\llav{\mathbf{Z}^{(m)}}_{m=1}^M$ spaces into a global $\G$ using specific linear transformations $\Wm$, that is, an FA model.

\begin{figure}
    \centering
    \includegraphics[width=0.3\linewidth]{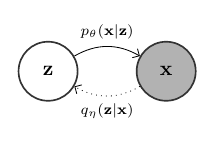}
    \caption{\ac{VAE} basic structure, where $q_\eta(\mathbf{z}| \mathbf{x})$ is the encoder network
    and $p_{\theta}(\mathbf{z}| \mathbf{x})$ is the decoder network. Grey circles denote observations, and white circles represent \ac{rv}.
    }
    \label{fig:vanillavae}
\end{figure}

The complete pipeline proposed in this work interconnects both previous solutions. To do so, each marginal \ac{VAE} is a probabilistic generative model that assumes that there exists a hidden embedded variable $\mathbf{Z}^{(m)}\in \R^{N \times D'}$ capable of generating the observations $\mathbf{X}^{(m)}\in\R^{N \times D}$ through a non-linear parametric model. Here, $D$ denotes the dimension of the observed data, and $D'$ is a hyperparameter for the Gaussian-embedded space. These embedded variables are inferred from the observations following:
\begin{equation}\label{eq:vaenonfeasible}
    p_{\eta}(\mathbf{Z}^{(m)}|\mathbf{X}^{(m)})= \frac{p_{\theta}(\mathbf{X}^{(m)}|\mathbf{Z}^{(m)})p(\mathbf{Z}^{(m)})}{p(\mathbf{X}^{(m)})},
\end{equation}
where, $p_{\theta}(\mathbf{X}^{(m)}| \mathbf{Z}^{(m)})$ corresponds to the generative decoder, which is a parametric model with $\theta$ parameters, and $p_\eta(\mathbf{Z}^{(m)}|\mathbf{X}^{(m)})$, with $\eta$ parameters, is the inference encoder as seen in Fig. \ref{fig:vanillavae}. 

A naive FA seeks to identify the global latent variables $\G$ that capture the covariance structure of a set of observed variables. In our framework, we redefine FA to identify global patterns from $m$-Gaussian-embedded spaces $\mathbf{Z}^{(m)}$. Specifically, we assume that each modality latent space sample $\Fnm \in \R^{1\times D'}$ can be generated as a linear combination of a smaller set of global factors, denoted as $\Gn\in \R^{1\times K}$, where $K<<D$, plus some added noise, $\mathbf{\epsilon}$. That is,
\begin{equation}
    \Fnm = \Gn\WmT + \mathbf{\epsilon} \,.
\end{equation}
where $\Wm\in \R^{D\times K}$ contains the coefficients that describe the relationship between each embedded variable $\Fnm$ and the global latent variable $\Gn$. Therefore, $\Wm$ selects which latent $K$ features of $\Gn$ are important to generate each specific $\Fnm$ modality.

\begin{figure}
\centering
\includegraphics[width=0.8\linewidth]{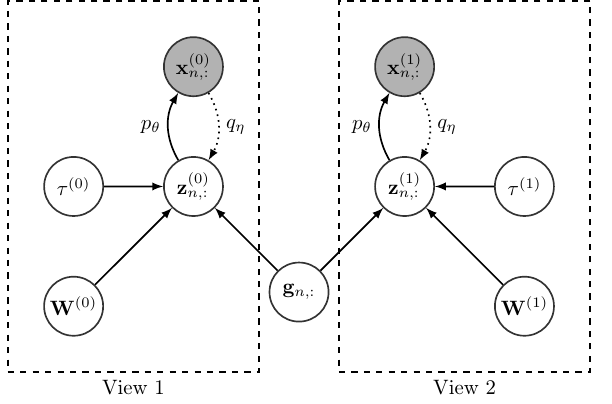}
\caption{FA-VAE graphical model example with two VAEs. Grey circles denote observations, and white circles represent \ac{rv}.}  \label{fig:relatedwork_faVAE}
\end{figure}

Hence, we propose a novel hierarchical approach to address the challenge of multi-modal analysis on heterogeneous datasets. FA-VAE uses marginal VAEs to transform the original multi-view data $\mathbf{X}^{(m)}$ into $m$-Gaussian latent spaces $\mathbf{Z}^{(m)}$. These latent spaces are then linearly combined with linear matrices $\Wm$ to create a global latent space $\G$, which captures the information shared between all views, as illustrated in Figure \ref{fig:relatedwork_faVAE}. The global latent variables are assumed to be isotropic:
\begin{align}
        & \Gn \sim \N\left(0, I_K\right) \,, 
\end{align}
as well as the linear projection matrices
\begin{align}
    & \Wkm \sim \N\left(0, \akm^{-1}I_K\right)  \,,
\end{align}
where, inheriting the philosophy of Bayesian Inter-Battery Factor Analysis (BIBFA) \cite{klami2013bayesian} and \ac{SSHIBA} \cite{sevilla-salcedo_sparse_2021} models, an $\akm$ term is added following a Gamma distribution to induce \ac{ARD}
\begin{align}
     & \akm \sim \Gamma\left(a^{\am}, b^{\am}\right) \,.
\end{align}
Therefore, each $k$-column of $\Wm$, $\Wkm$, follows a Gaussian distribution with zero mean and precision $\akm$. For this $\akm$ we assume a Gamma prior distribution allowing the precision to achieve high values. When this occurs, i.e., $\akm$ value increase significantly, all elements of $\Wkm$ tends to 0. This means that the $k$-th component of $\G$ do not affect to the generation of $\mathbf{Z}^{(m)}$, thus obtaining an automatic latent feature selection for each $m$ modality as they are constructed following:
\begin{align}\label{eq:embeddedlatentspace}
    & \mathbf{z}_{n,:}^{(m)} | \Gn \sim \N(\Gn\WmT, \taum^{-1}I_K)\,, 
\end{align}
where $\taum \sim \Gamma\left(a^{\taum}, b^{\taum}\right)$ is the precision of the noise parameter. 

For a given $m$-view, each marginal $m$-VAE infers a embedded Gaussian variable $\mathbf{z}_{n,:}^{(m)}$ via \ac{VI} from the observations following
\begin{equation}\label{eq:sampleVAE}
    \Fnm|\Onm \sim q_\eta(\Fnm) \sim \N\left(\mu^{(m)}_\eta(\Xnm), \Sigma^{(m)}_\eta(\Xnm)\right),
\end{equation}
where $\mu^{(m)}_\eta(\Xnm), \Sigma^{(m)}_\eta(\Xnm)$ are the output of an independent parametric encoder for the $m$-th view. Then, each marginal $m$-VAE generates the observation given the embedded $\Fnm$ using the conditional distribution $\Onm | \Fnm \sim p_\theta(\Onm|\Fnm)$, which is defined as
\begin{equation}
    \Onm | \Fnm \sim \N\left(\mu^{\Onm}_\theta(\Fnm), \sigma\right)
\end{equation}
where $\mu^{\Onm}_\theta(\Fnm)$ is the output of an independent parametric decoder for each $m$ view with $\sigma$ fixed. 

The FA-VAE architecture can handle various heterogeneous observations, such as images modelled using CNNs or sequential data modelled using RNNs, due to the flexibility in the encoder-decoder pairs of each $m$-VAE. This feature enables the FA-VAE model to effectively address the challenges posed by diverse and complex data types.

Employing vanilla VAEs for each $m$-VAE is a simplistic strategy that does not guarantee that the embedded spaces $\mathbf{Z}^{(m)}$ contain shared information, leading to a lack of correlation captured by $\G$. To address this limitation and ensure that the latent spaces do share information, we propose modifying the regularisation term of the vanilla VAE. In a vanilla VAE, the maximisation of the ELBO ensures that the variational distribution $q_\eta(\Fnm|\Onm)$ is close to the prior distribution $p(\Fnm)$ by minimising their KL (\textbf{II} term in \eqref{eq:vae_elbo}) divergence, following:
\begin{equation}\label{eq:vae_elbo}
    \mathcal{L}_{\theta, \eta} = \E_{q_{\eta}(\mathbf{z}_{n,:}^{(m)} \Xnm)} [\log\big(p_{\theta}(\Xnm| \mathbf{z}_{n,:}^{(m)})\big)]
    - \underbrace{\text{KL}\big(q_{\eta}(\mathbf{z}_{n,:}^{(m)}| \Xnm) | | p(\mathbf{z}_{n,:}^{(m)})\big)}_{\textbf{II}}  \,,
\end{equation}
In a standard VAE, the prior distribution $p(\mathbf{z}_{n,:}^{(m)})$ is typically assumed to be normally distributed, which results in a latent space that is both continuous and complete. However, there is no guarantee that the latent space provides meaningful shared information with other multimodal views. Thus, we propose using a prior distribution $p(\Fnm)$ that follows the equation given in Eq. \eqref{eq:embeddedlatentspace}, i.e. the posterior distribution assumed by the FA framework. Hence, the ELBO is constructed as follows:
\begin{equation}\label{eq:possible_elbo}
    \begin{split}
    \mathcal{L}^{(m)} &= \E_{q_\eta\left(\mathbf{\Fnm}| \Onm\right)} \log\left(p_\theta(\Onm| \mathbf{\Fnm})\right)\\
    &- \beta\text{KL}\left(q_\eta(\mathbf{\Fnm}|\Onm) || \N\left(\Gn\Wm, \taum^{-1}\right)\right),
    \end{split}
\end{equation}
where the first term is the \ac{GLL} between $\Onm$ and samples from $p_\theta(\Onm| \mathbf{\Fnm})$, and the second term minimises the KL divergence between the variational distribution and the posterior that the FA framework assumes which is a linear combination of the global latent variables and the projection matrices. The hyperparameter $\beta$ serves as a balance between two conflicting objectives: (i) the degree of fidelity in the reconstructed data and (ii) maximising the amount of shared information with the other views in the multi-modal dataset.

In terms of the optimisation process for the model's random variables, each $m$-latent variable $\mathbf{Z}^{(m)}$ is updated through its corresponding parametric decoder, while the remaining random variables in $\Theta$ are optimised using a mean-field approach. This involves iteratively updating each rv while holding the others fixed at their expected values. The specific update distribution rules for each rv are described in Table \ref{tab:multivae_up}.

\begin{table}[ht]
\caption{Distribution for all $\Theta$ obtained for the Multi-VAE Factor Analysis where $f_\eta^{(m)}(\cdot)$ is the marginal decoder $m$.}
\begin{adjustbox}{max width=\textwidth}
\renewcommand{\arraystretch}{1.}
\centering
\begin{tabular}{ccc}
\toprule
{\textbf{Variable}} & { $\bm{q}^*$ \textbf{distribution}} & \textbf{Parameters} \\\midrule
\multirow{3}{*}{$\Gn$}
& \multirow{3}{*}{$\N\big(\Gn | \mu_{\Gn},\Sigma_{\G}\big)$}
& $\mu_{\Gn} = \summ\bigg({\ang{\taum} \ang{\mathbf{Z}^{(m)}} \ang{\Wm} \Sigma_{\G}}\bigg)$ \\
& & $\Sigma_{\G}^{-1} = {I_{K} + \summ\bigg({\ang{\taum} \ang{\WmT \Wm}}}\bigg)$ \\&&\\
\midrule     
\multirow{3}{*}{$\mathbf{Z}^{(m)}$} &
\multirow{2}{*}{$\N\left(\mu^{(m)}_\eta(\Xnm), \Sigma^{(m)}_\eta(\Xnm)\right)$} 
& $\mu^{(m)}_\eta(\Xnm) = f_\eta^{(m)}(\Xnm) $\\&&$\Sigma^{(m)}_\eta(\Xnm) = f_\eta^{(m)}(\Xnm)$ \\ &&   \\
\midrule
\multirow{3}{*}{$\Wm$} 
& \multirow{2}{*}{$\prodd {\N \big({\Wdm | \mu_{\Wdm}, \Sigma_{\Wm}}}\big)$} 
& $\mu_{\Wdm} = \ang{\taum} \ang{\mathbf{Z}^{(m)}}^T \ang{\G} \Sigma_{\Wm}$ \\
& & $\Sigma_{\Wm}^{-1} = {\text{diag}(\ang{\am})  + \ang{\taum}\ang{\GT \G}}$  \\&&\\
\midrule    
\multirow{3}{*}{$\am$}
& \multirow{2}{*}{$\prodk \Gamma\big({\akm | a_{\akm},b_{\akm}}\big)$}
& $a_{\akm} = \frac{D_m}{2} + a^{\am}$ \\ 
& & $b_{\akm} = b^{\am} + \frac{1}{2} \sumd
 \ang{\Wdkm\Wdkm}$ \\&&\\
\midrule
\multirow{5}{*}{{$\taum$}}
& \multirow{5}{*}{$\Gamma\big({\taum | a_{\taum},b_{\taum}}\big)$}
& $a_{\taum} = \frac{D_mN}{2} + a^{\taum}$ \\
& & $b_{\taum} = b^{\taum} + \frac{1}{2} \left(\sumn\sumd \Xndm^2 \right.$ \\
& & $ - 2 \Tr\big({\ang{\Wm}\ang{\GT}\ang{\mathbf{Z}^{(m)}}}\big)$  \\
& & $ + \Tr\big({\ang{\WmT\Wm} \ang{\GT \G}}\big)\bigg)$  \\
\bottomrule
\end{tabular}
\end{adjustbox}
\label{tab:multivae_up}
\end{table}

\subsubsection{Binary or categorical views}\label{sec:bin_cat}

The previous FA-VAE is over-parameterised when one view is a category or a binary indicator. For such a case, a Bayesian Logistic Regression (BLR) model \cite{jaakkola1997variational} would be enough to translate from a binary observation, denoted as $\Xnm$, to its Gaussian distributed version $\Fnm$. 

Therefore, following SSHIBA \cite{sevilla-salcedo_sparse_2021}, when a binary or multi-label indicator is present
\begin{equation}
    p\p*{\Fnm|\Xnm} = \prodd{p\p*{z_{n,d}^{(m)} | x_{n,d}^{(m)}} }
\end{equation}
\begin{equation}
    p\p*{z_{n,d}^{(m)} | x_{n,d}^{(m)}} = e^{x_{n,d}^{(m)} z_{n,d}^{(m)}} \sigma \p*{-x_{n,d}^{(m)}}
\end{equation}
where for each multi-label view $\Fnm$ is now updated following mean-field as no VAE is used:
\begin{align}
    \Fnm &\sim \N \p*{\Fnm | \mu_{\Fnm }, \Sigma_{\Fnm}} \\
    \mu_{\Fnm } &= \Xnm - \frac{1}{2} + \ang{\taum}\ang{\Gn}\ang{\WmT}\\
    \Sigma_{\Fnm} &= \ang{\taum}I + 2 \Lambda_{\xi_{n,:}^{(m)}}
\end{align}

Therefore, two approximations to infer $\mathbf{Z}^{(m)}$ are proposed, as seen in Table \ref{tab:favae_up}. For complex $m$-views, that is, complex data types such as images, sequential or high-dimensional data, a $m$-VAE approach is taken following the first row of Table \ref{tab:favae_up} which corresponds to the FA-VAE approach. On the contrary, when dealing with a categorical or binary view, a simpler approach is taken to avoid overfitting and overparameterisation following the second row of Table \ref{tab:favae_up}.

\begin{table}[h]
\caption{$m$-Gaussian embedded space. The first row indicates how the embedded space is inferred when FA-VAE is proposed. Second row indicated how the embedded space is inferred when FA-VAE is proposed.}
\begin{adjustbox}{max width=\textwidth}
\renewcommand{\arraystretch}{1.}
\centering
\begin{tabular}{ccc}
\toprule
{\textbf{Variable}} & { $\bm{q}^*$ \textbf{distribution}} & \textbf{Parameters} \\\midrule
\multirow{3}{*}{$\mathbf{Z}^{(m)}$} &
\multirow{2}{*}{$\N\left(\mu^{(m)}_\eta(\Xnm), \Sigma^{(m)}_\eta(\Xnm)\right)$} 
& $\mu^{(m)}_\eta(\Xnm) = f_\eta^{(m)}(\Xnm) $\\&&$\Sigma^{(m)}_\eta(\Xnm) = f_\eta^{(m)}(\Xnm)$ \\ &&   \\
\midrule
\multirow{3}{*}{$\mathbf{Z}^{(m)}$} &
\multirow{2}{*}{$\N \p*{\Fnm | \mu_{\Fnm }, \Sigma_{\Fnm}}$} 
&  $\mu_{\Fnm } = \Xnm - \frac{1}{2} + \ang{\taum}\ang{\Gn}\ang{\WmT} $\\&&$\Sigma_{\Fnm} = \ang{\taum}I + 2 \Lambda_{\xi_{n,:}^{(m)}}$ \\ &&   \\
\bottomrule
\end{tabular}
\end{adjustbox}
\label{tab:favae_up}
\end{table}

\newpage

Following this procedure, FA-VAE trains according to the steps detailed in Algorithm \ref{alg:two}. For each particular $m$-view it updates the $\Fnm$ embedded variables by sampling from its distribution depending on whether is given by a VAE or by a BLR model.

\SetKwHangingKw{Init}{+}

\begin{algorithm}
\caption{FA-VAE training algorithm.}\label{alg:two}

\Init{$\G, \Wm, \Fnm, \akm, \taum$}
\While{FA-VAE not converge}{
Update $q(\G)$ following 1st row of Table \ref{tab:multivae_up}\;
\For{each $m$-view}{
Update $q(\Wm)$ following 2nd row of Table \ref{tab:multivae_up}\;
\eIf {$m$-view is not binary nor categorical}{
    \For{each epoch}{Maximise the $m$-VAE's ELBO (Eq. \eqref{eq:possible_elbo} )\;}
    Update $q(\Fnm)$ sampling from $q_\eta(\Fnm) \sim \N(\mu^{\Fnm}_\eta, \Sigma^{\Fnm}_\eta)$ \;
}{
 Update $q(\Fnm)$ following 2nd row of Table \ref{tab:favae_up}
}
Update $q(\akm)$ following 3th row of Table \ref{tab:multivae_up}\;
Update $q(\taum)$ following 4th row of Table \ref{tab:multivae_up}\;
}
}
\end{algorithm}

\clearpage

\section{Experiments}\label{sec:experiments}
Throughout this section, we demonstrate the flexibility of FA-VAE to address relevant problems in deep probabilistic modelling. In Section \ref{sec:first_scenario}, we analyse how FA-VAE can condition a pre-trained VAE to multi-label targets. In Section \ref{sec:dom_adap}, we apply FA-VAE to a domain adaptation problem and compare it with the Multi-VAE model \cite{Xu_2021_ICCV}. In addition, we disentangle and analyse the shared latent variables. Finally, in Section \ref{sec:trans_l} we use the proposed FA-VAE framework to perform transfer learning between multiple VAEs, showing how transfer learning creates a more expressive and understandable latent space than other models such as $\beta$-VAE \cite{burgess2018understanding}. The code to reproduce the following experiments can be found on GitHub\footnote{\url{https://github.com/aguerrerolopez/FA-VAE}}.

\subsection{FA-VAE as a conditioned generative model}\label{sec:first_scenario}
In this experiment, we demonstrate the utility of the FA-VAE framework in adapting a pre-trained unconditioned VAE, $p_\theta(\mathbf{x})$, to model a conditional distribution, $p_\theta(\mathbf{x}|\mathbf{a})$, using a given set of labelled data $\{\mathbf{x}_i, \mathbf{a}_i\}_{i=1}^{N}$, where $\mathbf{x}_i$ denotes an image and $\mathbf{a}_i$ denotes an attribute. We use the CelebA dataset \cite{liu2015faceattributes}, which consists of $30000$ samples of celebrity faces. Specifically, we selected three attributes, namely wearing lipstick, sex, and smiling. Each face is a $64 \times 64 \times 3$ RBG image, and the attributes are represented in a stratified manner throughout the dataset. 

\begin{figure}[h]
    \centering
    \includegraphics[width=0.6\linewidth]{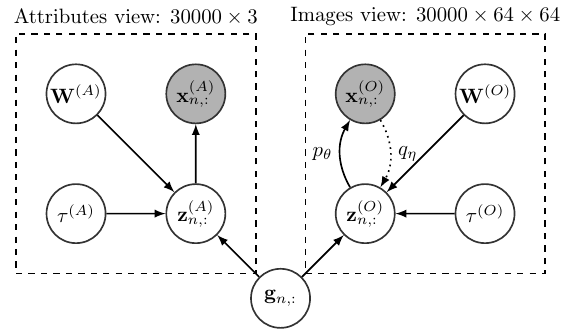}
    \caption{Conditioning a single VAE to a multi-label attribute vector using FA-VAE architecture where $A$ denotes attributes view and $O$ observations views. Gray circles are observations, and white circles represent \ac{rv}.}
    \label{fig:escenario_1_modelo}
\end{figure}

The graphical model is shown in Fig. \ref{fig:escenario_1_modelo}. The first view is responsible for modelling the attributes ($A$) as a multi-label vector with three attributes: wearing lipstick, gender, and smiling. For this view, we do not propose a VAE module, as it would be over-parameterised as explained in Section \ref{sec:bin_cat}. In this case, we generate the binary attributes denoted as $\mathbf{x}^{(A)}_{n,:}$ from the pseudo-observation $\mathbf{z}_{n,:}^{(A)}$ using a Bernoulli distribution. The mathematical expression for this distribution is given by
\begin{equation}
\begin{aligned}
p\big({\mathbf{x}^{(A)}_{n,:}|\mathbf{z}_{n,:}^{(A)}}\big) = \prod_{d=1}^{3} e^{z_{n,d}^{(A)} x_{n,d}^{(A)}}\sigma \bigg(-z_{n,d}^{(A)}\bigg) ,,
\end{aligned}
\end{equation}
where $\sigma$ denotes the sigmoid function.

In the second view ($O$), we consider RGB face images denoted as $\mathbf{x}^{(O)}_{n,:}\in \R^{64\times64\times3}$. For the encoder and decoder, we adopt the network architecture proposed in $\beta$-VAE paper \cite{higgins2016beta}. Specifically, we use a \acs{ CNN} with five convolutional layers, where the number of channels in each layer is $64,128,256,512,1024$, the size of the kernel is $4$, the stride is $2$ and the padding is $1$. Then, a fully connected layer is used to generate the parameters $\mu_\eta$ and $\Sigma_\eta$ to infer the embedded $\mathbf{z}_{n,:}^{O}\in\R^{100}$. The decoder network follows the inverse structure of the encoder.

\begin{figure}[h]
\centering
\begin{subfigure}{.45\textwidth}
  \centering
  \includegraphics[width=\linewidth]{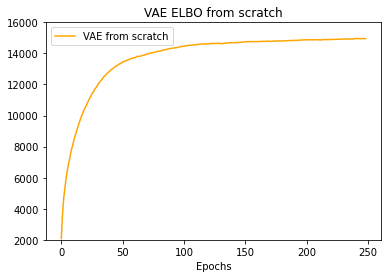}
  \caption{Vanilla VAE ELBO trained from scratch}
  \label{fig:escenario1_global}
\end{subfigure}%
\begin{subfigure}{.45\textwidth}
  \centering
  \includegraphics[width=\linewidth]{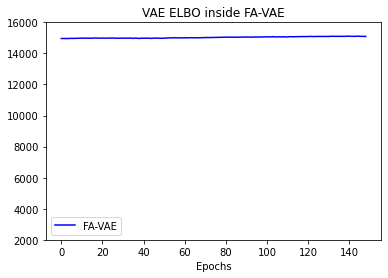}
  \caption{Pre-trained vanilla VAE ELBO inside FA-VAE's architecture}
  \label{fig:escenario1_recloss}
\end{subfigure}
\caption{VAEs convergence. Fig. \ref{fig:escenario1_global} shows the ELBO of the unsupervised VAE trained over CelebA from scratch. In Fig. \ref{fig:escenario1_recloss} we plug the unsupervised VAE from Fig. \ref{fig:escenario1_global} inside FA-VAE's architecture to condition it.}
\label{fig:escenario1_elbo}
\end{figure}

Initially, an unconditioned VAE is trained in a fully unsupervised manner on CelebA data until convergence. This pre-trained VAE is subsequently incorporated as the second view of FA-VAE to condition it on the attributes explained previously. The convergence of these two steps is analysed in Figure \ref{fig:escenario1_elbo}. Specifically, Figure \ref{fig:escenario1_global} shows the evolution of ELBO of unsupervised VAE trained from scratch, while Figure \ref{fig:escenario1_recloss} portrays the progression of ELBO during FA-VAE fine-tuning. These findings indicate that the pre-trained VAE remains stable and does not lose reconstruction power while being conditioned on three attributes.

\begin{figure}[h]
\centering
\begin{subfigure}{.5\textwidth}
  \centering
  \includegraphics[width=\linewidth]{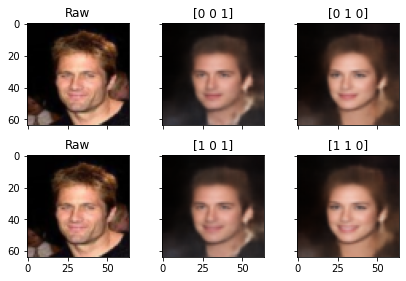}
  \caption{Example 1}
  \label{fig:escenario1_modface1}
\end{subfigure}%
\begin{subfigure}{.5\textwidth}
  \centering
  \includegraphics[width=\linewidth]{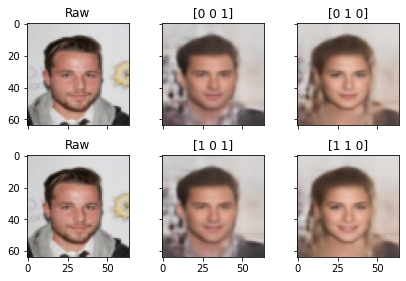}
  \caption{Example 2}
  \label{fig:escenario1_modface2}
\end{subfigure}
\caption{Different faces are generated by FA-VAE when modifying their attributes. The left column of each subfigure represents the raw image. Each subfigure's centre and right columns represent the altered images by changing the different attributes indicated in the title, meaning [smile, lipstick, gender].}
\label{fig:escenario1_facesmod}
\end{figure}

We can also modify the attributes of a given face. The resulting images are presented in Figs. \ref{fig:escenario1_modface1} and \ref{fig:escenario1_modface2}. These images demonstrate the ability of our model to alter attributes of the input image, such as gender (right column) or facial expression (smiling or neutral, top row vs. bottom row). We can also generate random images conditioned to any arbitrary set of attributes, as can be seen in \ref{sec:appendix_favaecond}.

\newpage

\subsection{Domain adaptation}\label{sec:dom_adap}
In this experiment, we explore the modularity of FA-VAE and demonstrate its ability to combine multiple VAEs simultaneously. To illustrate this, we employ a three-view FA-VAE configuration, using the CelebA dataset \cite{liu2015faceattributes} and the Google Cartoon Set dataset \cite{royer2020xgan}. Specifically, we train a VAE with $10000$ CelebA images in the first view and another VAE with $10000$ Cartoon images in the second view.

\begin{figure}[h]
    \centering
    \includegraphics[width=\linewidth]{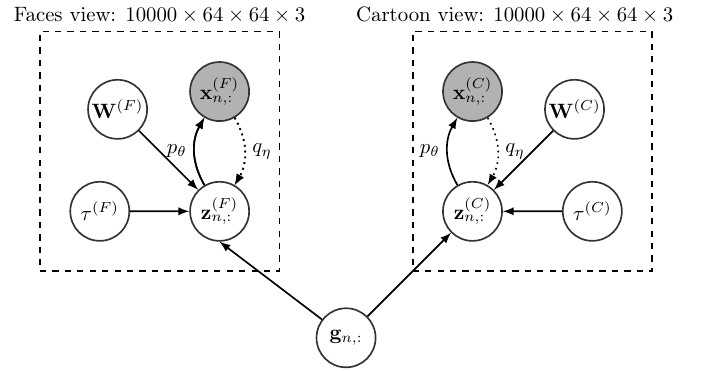}
    \caption{FA-VAE configuration to perform domain adaptation between two VAE-based views representing real-world faces and cartoon avatars. Grey circles denote observations, and white circles represent \ac{rv}.}
    \label{fig:esc3graph}
\end{figure}

Our hypothesis is that each $m$-embedded variable $\mathbf{z}_{n,:}^{(m)}$ can capture domain information from face images and cartoon avatars, respectively. The shared variable in latent space, $\Gn$, serves as a bridge between domains, facilitating the adaptation of real-world faces to 2D cartoon avatars. The architecture is depicted in Fig. \ref{fig:esc3graph}. This approach has potential applications in various domains, including face recognition, character creation, and animation.

The CelebA face view ($F$) comprises $10000\times64\times64\times3$ real celebrity face images, while the Cartoon view ($C$) consists of $10000\times64\times64\times3$ cartoon avatar images. We use the $\beta$-VAE encoder-decoder configuration proposed in Section \ref{sec:first_scenario} by \cite{burgess2018understanding} with $\beta>1$ for both views.

\begin{figure}[h]
    \centering
    \includegraphics[width=0.4\linewidth]{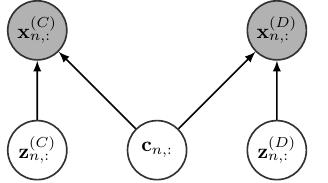}
    \caption{Conditiong two VAEs with a categorical variable using Multi-VAE model. CelebA images are represented by $\mathbf{x}_{n,:}^{(F)}$, Cartoon images are represented by $\mathbf{x}_{n,:}^{(C)}$, and the $\mathbf{c}_{n,:}$ captures the shared information. Grey circles denote observations, and white circles represent \ac{rv}.}
    \label{fig:relatedwork_MultiVAE}
\end{figure}

We conducted an experiment comparing FA-VAE with the Multi-VAE \cite{Xu_2021_ICCV} model, which uses a discrete latent variable $\mathbf{c}_{n}$ to share the context of all views (see Fig. \ref{fig:relatedwork_MultiVAE}).

\begin{figure}[h]
    \centering
    \includegraphics[width=\linewidth]{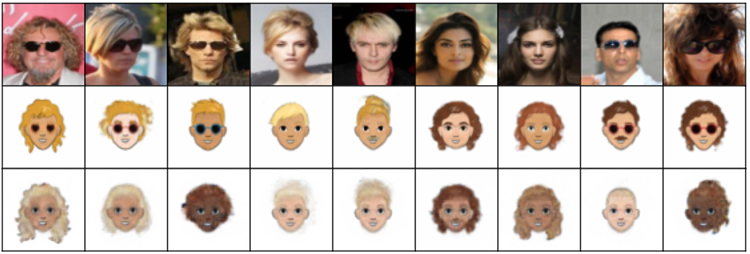}
    \caption{Domain adaptation from CelebA dataset to Cartoon dataset. The first row represents the original observations in CelebA dataset. In contrast, the second and third rows represent their translation to the Cartoon domain using \textbf{FA-VAE} (second row) and \textbf{Multi-VAE} (third row).}
    \label{fig:esc3_domainadap}
\end{figure}

Fig. \ref{fig:esc3_domainadap} presents a comparison of the performance of the FA-VAE and Multi-VAE models in the task of translating images from the CelebA domain to the Cartoon domain. The first row shows CelebA images, the second row displays the generated images by FA-VAE, and the third row shows the generated images by Multi-VAE. FA-VAE outperforms Multi-VAE in capturing inherent features such as sunglasses, in addition to hair colour. Multi-VAE fails to properly learn skin colour, as evidenced by Images 3 and 9, while FA-VAE produces high-quality 2D avatars without blur or artefacts.

\begin{figure}[h]
    \centering
    \includegraphics[width=\linewidth]{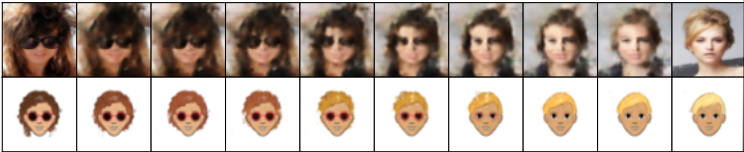}
    \caption{Examples of an image transformation and domain adaption application. The first and second rows show the evolution from $\mathbf{z}^{(F)}_{1,:}$ to $\mathbf{z}^{(F)}_{2,:}$ and, from $\mathbf{z}^{(C)}_{1,:}$ to $\mathbf{z}^{(C)}_{2,:}$, respectively.}
    \label{fig:esc3_ztrans1}
\end{figure}

The embedded latent variables $\mathbf{Z}^{(F)}$ and $\mathbf{Z}^{(C)}$ codify the information of each domain, while the global space $\G$ allows the transference of information between both domains. We can use this to apply domain adaption by translating images from one domain to the other. For this purpose, we first obtain two CelebA images and project them into their embedded latent space as $\mathbf{z}^{(F)}_{1,:}$ and $\mathbf{z}^{(F)}_{2,:}$. Then, we project them onto the shared latent space $\mathbf{g}_{1,:}, \mathbf{g}_{2,:}$ to, later, sample from the cartoon embedded space obtaining $\mathbf{z}^{(C)}_{1,:}$ and $\mathbf{z}^{(C)}_{2,:}$. Now, we interpolate the embedded representation in each domain using the convex combination
\begin{equation}\label{eq:esc3_ztrans1}
    \Fnm_{\lambda,:} = \lambda\Fnm_{1,:} + (1-\lambda)\Fnm_{2,:},
\end{equation}
where $m\in[C,F]$ represents any of the two views, and $\lambda \in [0,1]$ allows us to move from Image 1 to Image 2. Finally, using the embedded representation generated $\mathbf{z}^{(m)}_{\lambda,:}$ we can use $p^{(m)}_{\theta}(\mathbf{x}^{(m)}_{\lambda,:}|\mathbf{z}^{(m)}_{\lambda,:})$ to generate the sample for each domain. The reconstructed sequences $\mathbf{x}^{(m)}_{\lambda,:}|\mathbf{z}^{(m)}_{\lambda,:}$ for different values of $\lambda$ are shown in Fig. \ref{fig:esc3_ztrans1}.

Fig. \ref{fig:esc3_ztrans1} shows two rows of images showing the complete evolution from one image to the other through points sampled from the embedded space. The completeness of the embedded variables is demonstrated by the generation of meaningful images from any sampled point. Additionally, the gradual and smooth transitions between images demonstrate the continuity of the embedded variables. In the cartoon domain, the images show a clear evolution in hairstyle, hair colour, and eyeglasses. Therefore, we can conclude that both embedded variables, $\mathbf{Z}^{(C)}$ and $\mathbf{Z}^{(F)}$, are informative and explainable, satisfying both completeness and continuity criteria. Similarly, the global latent space $\G$ also demonstrates completeness and continuity, as shown in \ref{sec:appendix_domadap}.

\begin{figure}[h]
    \centering
    \includegraphics[width=\linewidth]{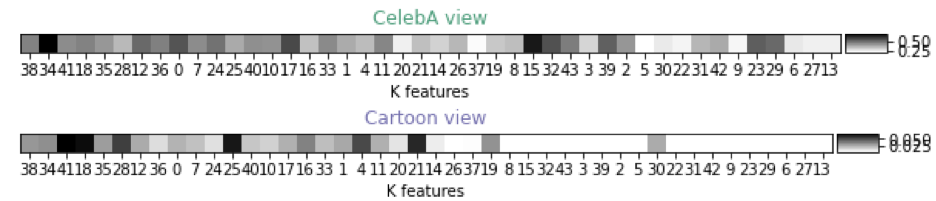}
    \caption{Mean over the rows of each $\Wm$ matrix. Each row is a vector representing the importance that each $k$ latent feature of $\Gn$ has to reconstruct each view. The first row represents the Celeba view, the second row represents the Cartoon view, and the third row represents the Hair view.}
    \label{fig:esc3_wmatrices}
\end{figure}

We can further analyse the interrelationships between multiple views of a dataset. To achieve this, we calculate the mean of each matrix $\Wm$ along its rows, resulting in a row vector of length $K$, where $K$ is the number of latent factors in $\Gn$. This vector represents the relative importance of each latent factor in $\Gn$ for each view. The two resulting vectors are shown in Fig. \ref{fig:esc3_wmatrices}, sorted by weight values in the hair view. The shared and unique latent features across views are then analysed on the basis of their corresponding weights in both views.

The results show that all latent factors are needed to explain CelebA faces, as they are a more complex domain. On the contrary, almost half of the latent factors are only relevant to CelebA faces and not to their cartoon version.

\subsection{Transfer learning}\label{sec:trans_l}
The computational cost of training powerful VAEs is a significant challenge, as these deep networks require many epochs to converge. To address this issue, we propose a novel approach to accelerate the training process using FA-VAE as a transfer learning tool between multiple \ac{VAE}s operating on the same domain.

\begin{figure}[h]
    \centering
    \includegraphics[width=0.8\linewidth]{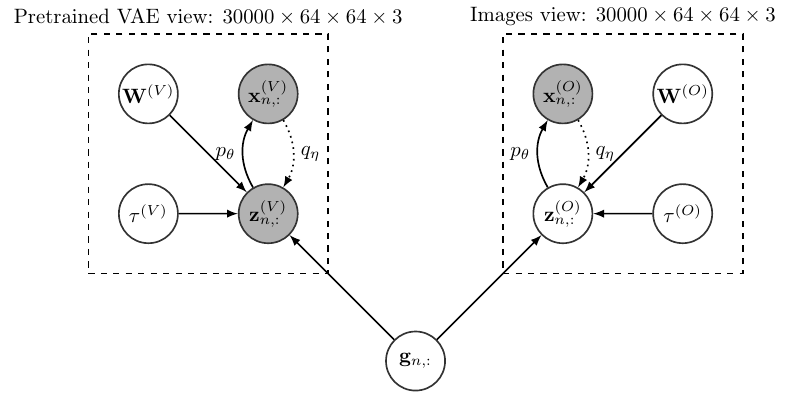}
    \caption{Transfer learning graphical model using FA-VAE. The $V$ view represents information pre-learned by a vanilla VAE. As it is pre-trained, $\mathbf{z}_{n,:}^{(V)}$ is no longer a \ac{rv} but an observation. The $O$ view represents CelebA images using a $\beta$-VAE. Grey circles denote observations, and white circles represent \ac{rv}.}
    \label{fig:escenario_2_modelo}
\end{figure}

Consider the CelebA dataset as an illustrative example, with a two-view setup described by the graphical model shown in Figure \ref{fig:escenario_2_modelo}. In the first view ($V$), we employ a vanilla pre-trained VAE. As such, we only use the pre-trained encoder and decoder and do not train them again, resulting in static $\mathbf{z}_{n,:}^{(V)}$. For the second view ($O$), we begin with the architecture of the $\beta$-VAE discussed in Section \ref{sec:first_scenario} and add an additional final CNN layer to make it deeper. The final CNN layer has a channel size of 2048, kernel size of 4, stride of 2, and padding of 1 in the encoder, while the decoder exhibits an inverse structure.

\begin{figure}[h]
\centering
\begin{subfigure}{.5\textwidth}
  \centering
  \includegraphics[width=\linewidth]{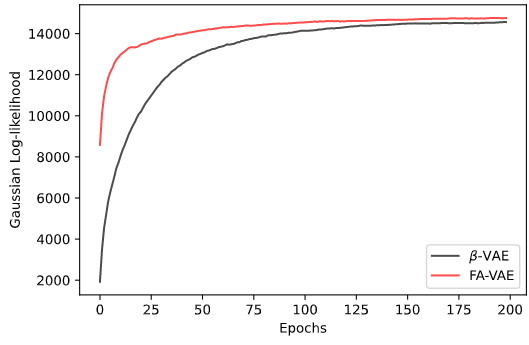}
  \caption{Reconstruction term measured by \ac{GLL}.}
  \label{fig:escenario2_recloss}
\end{subfigure}%
\begin{subfigure}{.5\textwidth}
  \centering
  \includegraphics[width=\linewidth]{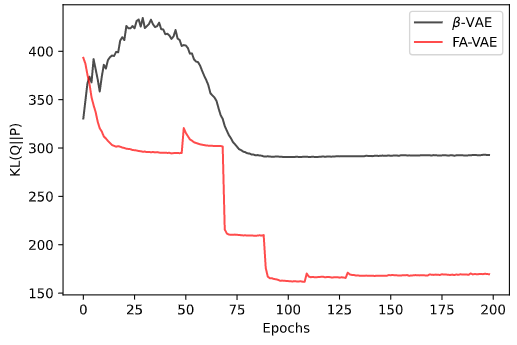}
  \caption{KL divergence term.}
  \label{fig:escenario2_kl}
\end{subfigure}
\caption{ELBO decomposition in reconstruction term and KL divergence term. The red line represents our approach, FA-VAE, while the black line represents the $\beta$-VAE on its own.}
\label{fig:escenario2_elbo}
\end{figure}

Our hypothesis is that using the embedded space provided by a pre-trained vanilla VAE, $\mathbf{z}_{n,:}^{(V)}$, we can accelerate the training of a deeper VAE and potentially lead to a better global solution. To demonstrate this hypothesis, we compared two scenarios: (i) two views FA-VAE with the vanilla VAE on view 1 and the $\beta$-VAE on view 2, i.e., transfer learning structure, and (ii) the $\beta$-VAE on its own. We present the performance of both approaches in Figure \ref{fig:escenario2_elbo}. We observe that FA-VAE demonstrates accurate reconstruction capabilities in the initial epochs, indicating that the latent space of the vanilla VAE provides a good initialisation. Furthermore, FA-VAE achieves the same maximum GLL as $\beta$-VAE in approximately four times fewer epochs, demonstrating its faster speed. Additionally, FA-VAE achieves the highest absolute value in terms of GLL compared to $\beta$-VAE alone. The global term of KL for FA-VAE is lower than that for $\beta$-VAE, as shown in Figure \ref{fig:escenario2_kl}. The periodic spikes in the KL plot correspond to the times when the SSHIBA part of FA-VAE updates the VAE prior distribution, as seen in Algorithm \ref{alg:two}. Meanwhile, for $\beta$-VAE, KL starts increasing while FA-VAE directly decreases, which justifies the better behaviour since the beginning.

\begin{figure}
    \centering
    \includegraphics[width=0.5\linewidth]{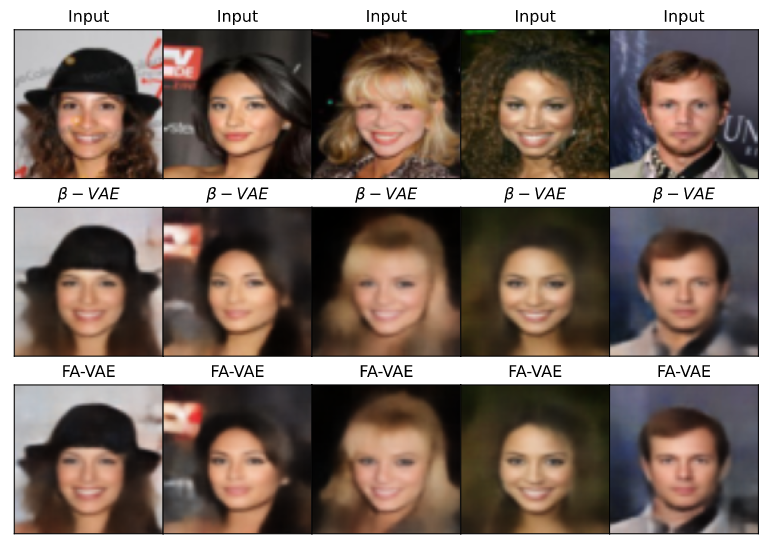}
    \caption{Images reconstructed by $\beta$-VAE and by FA-VAE}
    \label{fig:esc2_recimages}
\end{figure}

In Fig. \ref{fig:esc2_recimages}, we present five randomly selected test images that were not used during the training phase. We applied two different models, namely $\beta$-VAE and FA-VAE, to encode the images into their corresponding latent spaces, followed by reconstructing them back to their original domain. Since it might not be straightforward to discern the method that produces the best reconstruction, we provide the R2 score for the reconstruction of 10,000 test images in Table \ref{tab:esc2_r2_rec}. As can be seen from the table, FA-VAE outperforms the other models in terms of the R2 score, indicating that transfer learning can enhance model performance.

\begin{table}[]
    \centering
    \begin{tabular}{c c c}
    \toprule
         Model & Samples & R2 score \\\midrule
         Multi-VAE & 10000 & $0.855 \pm 0.154$  \\ 
         $\beta$-VAE & 10000 & $0.941 \pm 0.032$ \\ 
         FA-VAE & 10000 & $\mathbf{0.969 \pm 0.027}$  \\ 
    \bottomrule
    \end{tabular}
    \caption{Reconstruction performance measured in R2 score over 10,000 CelebA test samples.}
    \label{tab:esc2_r2_rec}
\end{table}

FA-VAE offers the advantage of creating a more expressive and meaningful embedded latent representation of images compared to $\beta$-VAE. To demonstrate it, the 10 most relevant features are arbitrarily modified to analyse their visual impact on the reconstructed image. The 10 most relevant characteristics were selected based on the absolute values in the weight matrix $\Wm$, for both embedded latent variables: $\mathbf{z}\in\R^{1x100}$ for $\beta$ -VAE and $\mathbf{z}_{n,:}^{(O)}\in\R^{1x100}$ for FA-VAE. These values were then randomly modified in the $[-20,20]$ interval.

\begin{figure}
    \centering
    \begin{subfigure}{0.48\textwidth}
        \includegraphics[width=\linewidth]{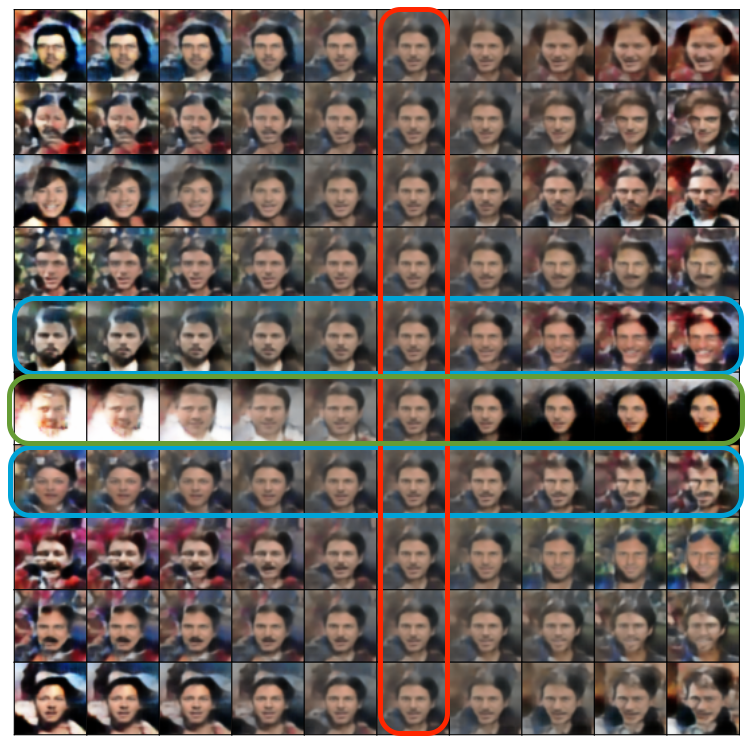}
        \caption{$\beta$-VAE}
        \label{fig:bvae_latent}
    \end{subfigure}
    ~
    \begin{subfigure}{0.48\textwidth}
        \includegraphics[width=\linewidth]{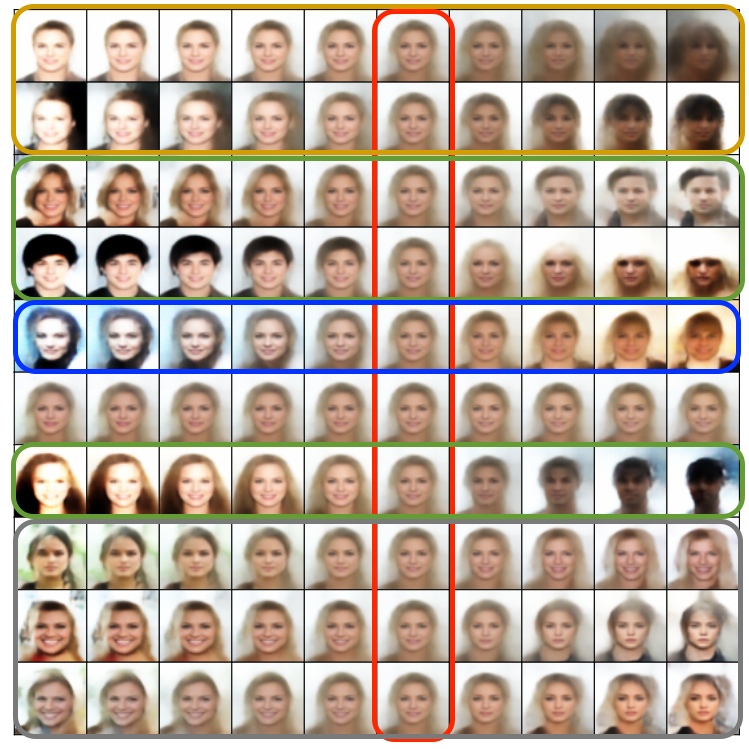}
        \caption{FA-VAE}
        \label{fig:img1}
    \end{subfigure}
    \caption{Latent space evolution. Each row represents the 10 most relevant features based on absolute value. The red column represents the image generated by the model without any modification. Then, all images at the left of the red column are images generated by arbitrarily decreasing the value of each feature. Likewise, all images at the right of the red column are images generated by increasing the value of each component.}
    \label{fig:latent}
\end{figure}

As shown in Fig. \ref{fig:bvae_latent}, for $\beta$-VAE, each row represents one of the 10 most relevant characteristics. The latent space $\mathbf{z}$ reveals that only three features have a visual interpretation. The blue rows show that facial hair can be controlled by increasing or decreasing its corresponding value. The green row also has an impact on the image contrast. However, the remaining features do not have a discernible visual impact on the images.

Figure \ref{fig:img1} presents an evaluation of FA-VAE that demonstrates its superiority over $\beta$-VAE in terms of both the interpretability and clarity of its embedded latent space. Among the 10 most relevant features identified, 9 are visually interpretable. For example, the golden-marked latent features control skin tone and facial rotation, the green rows determine gender and hairstyle, the blue row distinguishes between cold and warm background colours, and the grey-labelled latent features regulate smiling. Furthermore, the study shows that FA-VAE can effectively capture facial information while filtering out noisy backgrounds in the latent space.

\section{Conclusions}\label{sec:conc}
This paper introduces the FA-VAE model, which is a novel deep hierarchical VAE designed to handle mixed and heterogeneous data types, using a highly interpretable FA latent space. This model has the ability to condition multiple VAEs, enabling it to work with various data domains such as continuous, binary, categorical, and image data, depending on the VAE architecture. This is achieved through a deep hierarchical structure that utilises FA to learn a disentangled and explainable latent space.

Furthermore, we demonstrate the effectiveness of the model in domain adaptation between two different databases by conditioning them on external attributes. Additionally, FA-VAE is the first model capable of performing transfer learning between generative models, thus speeding up the learning process and improving performance. Overall, FA-VAE provides a robust and versatile model for dealing with real-world datasets, making it a valuable tool for various applications.

\section*{CRediT authorship contribution statement}
\textbf{Alejandro Guerrero-López}: Conceptualisation, Data curation, Formal Analysis, Investigation, Methodology, Software, Writing - original draft, Writing - review and \& editing.\textbf{ Carlos Sevilla-Salcedo}: Conceptualisation, Formal Analysis, Investigation, Methodology, Writing - original draft, Writing - review and \& editing.\textbf{ Vanessa Gómez-Verdejo}:  Project administration, Supervision, Resources, Funding acquisition, Writing - review \& editing \textbf{Pablo M. Olmos}: Project administration, Supervision, Resources, Funding acquisition, Writing - review \& editing

\section*{Declaration of Competing Interest}
The authors declare that they have no known competing financial interests or personal relationships that could have appeared to influence the work reported in this document.

\section*{Acknowledgements}
This work was supported by MCIN/AEI/10.13039/501100011 [grant numbers RTI2018-099655-B-100, PID2020-115363RB-I00,  PID2021-123182OB-I00, TED2021-131823B-I00]; the Comunidad de Madrid [grant numbers IND2018/TIC-9649, Y2018/TCS-4705]; the European Union (FEDER and the European Research Council (ERC) through the European Unions Horizon 2020 research innovation program [grant number 714161]; and the IISGM [grant INTRAMURAL].

\bibliography{paper-principal.bib}

\newpage
\appendix

\section{Experiments}
In this Appendix, we add different experiments that demonstrate the potential of FA-VAE. The structure follows the same order as in the main article.

\subsection{FA-VAE as a conditioned generative model} \label{sec:appendix_favaecond}
Following the experiment shown in Section \ref{sec:first_scenario}, we demonstrate that we can also generate random conditioned faces.

\begin{figure}[h]
\centering
\includegraphics[width=0.6\linewidth]{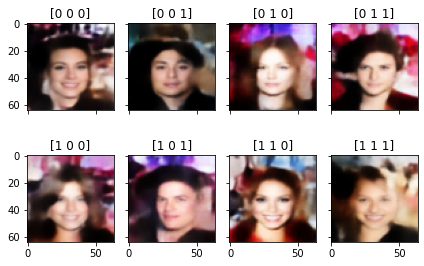}
\caption{Fake faces generated by random $\mathbf{x}^{(A)}_{n,:}$ vectors. The title of each image indicates which attribute is activated: smiling, wearing lipstick, and gender. For example, [1 0 0] means a smile [1] without lipstick [0] on a woman's face [0], and [1 0 1] means a smile [1] without lipstick [0] on a male's face [1].}
\label{fig:escenario1_faces}
\end{figure}

To analyse it, we show how the proposed model can generate images conditioned to specific attributes. We proceed as follows: (i) we create random $\mathbf{x}^{(A)}_{n,:}$ multi-label vectors using [wearing lipstick, gender, smiling] binary notation, (ii) we create their real pseudo-observation $\Xan$, (iii) we generate the posterior distribution of $\Gn$ given $\Xan$ which follows a Gaussian distribution with parameters:
\begin{equation}\label{eq:zpred}
    \begin{split}
    &\Sigma_{\mathbf{g}^{*}_{n,:}}^{-1}= I_{K_c} + \taua\Wa^T\Wa\\
    &\mu_{\mathbf{g}^{*}_{n,:}}= \taua\Xan\Wa\Sigma_{\mathbf{g}^{*}_{n,:}},
    \end{split}
\end{equation}
(iv) we sample $\Xon \sim \N(\mathbf{g}^{*}_{n,:}\Wo, \tauo)$, and, finally (v) we use FA-VAE's generative distribution $p_\theta(\mathbf{o}_{n,:}|\Xon)$ to sample artificially generated conditioned images. Fig. \ref{fig:escenario1_faces} shows these images generated by 8 random $\mathbf{x}^{(A)}_{n,:}$ multilabel attributes.

We can quickly condition a pre-trained VAE to arbitrary attributes by training 150 epochs of FA-VAE.

\subsection{Domain adaptation} \label{sec:appendix_domadap}
Following the experiment described in Section \ref{sec:dom_adap}, we show that the global latent variable is also complete and continuous.

\begin{figure}[h]
    \centering
    \includegraphics[width=\linewidth]{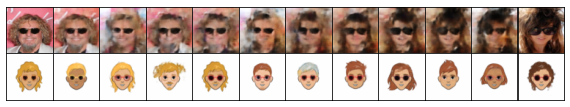}
    \caption{Example of an image transformation and domain adaptation application using a common global representation $\mathbf{g}_{\lambda,:}$. The first row shows images generated by the CelebA VAE while the second row shows images generated by the Cartoon VAE.}
    \label{fig:esc3_ztrans2}
\end{figure}

Similarly, as in \ref{sec:dom_adap} with the interpolation of private latent space, we could implement this interpolation directly from the global space by sampling from $\Gn$ and generating a sample that both domains can understand. The next two examples illustrate this behaviour. Again, we select two new CelebA observations and project them to the global space $\Z$ as $\mathbf{g}_{1,:}$ and $\mathbf{g}_{2,:}$. Then, we calculate $ \mathbf{g}_{\lambda,:}$ following Eq. (\ref{eq:esc3_ztrans1}). For this scenario, each generative network decodes $\mathbf{g}_{\lambda,:}$ creating a pair of images. In Fig. \ref{fig:esc3_ztrans2} the reconstructed sequences $\mathbf{x}^{(m)}_{\lambda,:}|\mathbf{g}_{\lambda,:}$ are shown.

Each of the $\mathbf{g}_{\lambda,:}$ generates meaningful content showing completeness. However, there is a trade-off between both domains. On the one hand, the CelebA domain shows better continuity, leading to a clear transition between both images. On the other hand, the Cartoon domain, characterised by a discrete set of features, shows better completeness where every $\mathbf{g}_{\lambda,:}$ creates a cartoon avatar without any distortion or artefact.

\end{document}

%% file: formulas.tex
\newcommand{\R}{\mathbb{R}} 
\newcommand{\E}{\mathbb{E}} 
\newcommand{\N}{\mathcal{N}} 
\DeclareMathOperator{\Tr}{Tr} 

\DeclarePairedDelimiter\p{(}{)}

\DeclarePairedDelimiter\llav{\{}{\}}
\DeclarePairedDelimiter\ang{\langle}{\rangle}

\DeclareMathOperator{\Xon}{\mathbf{X}_{n,:}^{(O)}} 
\DeclareMathOperator{\Xan}{\mathbf{X}_{n,:}^{(A)}} 
\DeclareMathOperator{\Xm}{\mathbf{X}^{(m)}} 
\DeclareMathOperator{\Xnm}{\mathbf{x}_{n,:}^{(m)}} 
\DeclareMathOperator{\Fnm}{\mathbf{z}_{n,:}^{(m)}} 
\DeclareMathOperator{\Onm}{\mathbf{x}_{n,:}^{(m)}} 
\DeclareMathOperator{\Xndm}{x_{n,d}^{(m)}} 

\DeclareMathOperator{\Z}{\mathbf{Z}} 

\DeclareMathOperator{\G}{\mathbf{G}} 
\DeclareMathOperator{\GT}{\mathbf{G}^T} 
\DeclareMathOperator{\Gn}{\mathbf{g}_{n,:}} 

\DeclareMathOperator{\Wo}{\mathbf{W}^{(O)}} 
\DeclareMathOperator{\Wa}{\mathbf{W}^{(A)}} 
\DeclareMathOperator{\Wm}{\mathbf{W}^{(m)}} 
\DeclareMathOperator{\WmT}{\mathbf{W}^{(m)^T}} 
\DeclareMathOperator{\Wkm}{\mathbf{w}_{:,k}^{(m)}} 
\DeclareMathOperator{\Wdm}{\mathbf{w}_{d,:}^{(m)}} 
\DeclareMathOperator{\Wdkm}{w_{d,k}^{(m)}} 

\DeclareMathOperator{\am}{\text{\boldmath$\alpha$}^{(m)}} 
\DeclareMathOperator{\akm}{\alpha_k^{(m)}} 

\DeclareMathOperator{\tauo}{\tau^{(O)}} 
\DeclareMathOperator{\taua}{\tau^{(A)}} 
\DeclareMathOperator{\taum}{\tau^{(m)}} 

\DeclareMathOperator{\sumd}{\sum\limits_{d=1}^{D_m}} 
\DeclareMathOperator{\sumn}{\sum\limits_{n=1}^{N}} 
\DeclareMathOperator{\summ}{\sum\limits_{m=1}^{M}} 

\DeclareMathOperator{\prodd}{\prod\limits_{d=1}^{D_m}} 
\DeclareMathOperator{\prodk}{\prod\limits_{k=1}^{K_c}} 

